\documentclass{ecai}
\usepackage{latexsym}

\usepackage{times}
\usepackage{soul}
\usepackage{url}
\usepackage[hidelinks]{hyperref}
\usepackage[utf8]{inputenc}
\usepackage[small]{caption}
\usepackage{graphicx}
\usepackage{amsmath}
\usepackage{booktabs}
\usepackage{algorithm}
\usepackage{algorithmic}
\usepackage{url,ifthen}
\usepackage{amssymb}

\usepackage{fancybox}
\usepackage{tikz}

\usepackage{grbbridge}
\usepackage{enumitem}

\ecaisubmission   

\title{The $\alpha\mu$ Search Algorithm for the Game of Bridge}

\author{Tristan Cazenave \institute{LAMSADE Université Paris-Dauphine PSL CNRS, France, email: Tristan.Cazenave@dauphine.psl.eu} \and Véronique Ventos \institute{NUKKAI, Paris, France, email: vventos@nukk.ai}}

\begin{document}

\maketitle

\begin{abstract}
$\alpha\mu$ is an anytime heuristic search algorithm for incomplete
information games that assumes perfect information for the
opponents. $\alpha\mu$ addresses the strategy fusion and non-locality
problems encountered by Perfect Information Monte Carlo sampling. In
this paper $\alpha\mu$ is applied to the game of Bridge.
\end{abstract}

\section{Introduction}

As superhuman level has been reached for Go starting from zero
knowledge \cite{silver2017mastering} and as it is also the case for
other two player complete information games such as Chess and
Shogi \cite{silver2018general} some of the next challenges in games
are imperfect information games such as Bridge or Poker. Multiplayer
Poker has been solved very recently \cite{brown2019superhuman} while
computer Bridge programs are still not superhuman.

The state of the art for computer Bridge is Perfect Information Monte
Carlo sampling (PIMC). It is a popular algorithm for imperfect
information games. It was first proposed by
Levy \cite{levy1989million} for Bridge, and used in the popular
program GIB \cite{ginsberg2001}. PIMC can be used in other
trick-taking card games such as
Skat \cite{buro2009improving,kupferschmid2006skat}, Spades and
Hearts \cite{sturtevant2006feature}. The best Bridge and Skat programs
use PIMC. Long analyzed the reasons why PIMC is successful in these
games \cite{long2010understanding}.

However PIMC plays sub-optimally due to two main problems: strategy
fusion and non locality. We will illustrate these problems in the
second section. Frank and Basin \cite{frank2001theoretical} have
proposed a heuristic algorithm to solve Bridge endgames that addresses
the problems of strategy fusion and non locality for late
endgames. The algorithm we propose is an improvement over the
algorithm of Frank and Basin since it solves exactly the endgames
instead of heuristically and since it can also be used in any state
even if the search does not have enough time to reach the terminal
states. Ginsberg has proposed to use a lattice and binary decision
diagrams to improve the approach of Frank and Basin for solving Bridge
endgames \cite{ginsberg2001}. He states that he was generally able to
solve 32 cards endings, but that the running times were increasing by
two orders of magnitude as each additional card was added. $\alpha\mu$
is also able to solve Bridge endings but it can also give a heuristic
answer at any time and for any number of cards and adding cards or
searching deeper does not increase as much the running time.

Furtak has proposed recursive Monte Carlo search for
Skat \cite{furtak2013recursive} to improve on PIMC but the algorithm
does not give exact results in the endgame and does not solve the non
locality problem.

Other approaches to imperfect information games are Information Set
Monte Carlo Tree Search \cite{cowling2012information}, counterfactual
regret minimization \cite{zinkevich2008regret}, and Exploitability
Descent \cite{lockhart2019computing}.

$\alpha\mu$ searches with partial orders. It is related to partial
order bounding \cite{muller2001partial} and to searching game trees
with vectors of integer values \cite{dasgupta1996searching}. However
our algorithm is different from these algorithms since it searches
over vectors only composed of 0 and 1 and uses different backups for
sets of vectors at Max and Min nodes as well as probabilities of
winning.

The contributions of the paper are:

\begin{enumerate}[topsep=0pt]
\item An anytime heuristic search algorithm that assumes Min
players have perfect information and that improves on PIMC and
previous related search algorithms.
\item An anytime solution to the strategy fusion problem of PIMC that solves the strategy fusion problem when given enough time.
\item An anytime solution to the non-locality problem of PIMC using Pareto
fronts of vectors representing the outcomes for the different possible
worlds. It also converges given enough time.
\item A search algorithm with Pareto fronts.
\item The description of the early and root cuts that speed up the search.
\item Adaptation of a transposition table to the algorithm so as to
improve the search speed using iterative deepening.
\item Experimental results for the game of Bridge.
\end{enumerate}

The paper is organized as follows: the second section deals with PIMC
for computer Bridge and its associated defects. The third section
defines vectors of outcomes and Pareto fronts. The fourth section
deals with search with strategy fusion and non locality. The fifth
section gives experimental results.

\section{Perfect Information Monte Carlo Sampling}

In this section we illustrate the problems of PIMC.

\subsection{Double Dummy Solver}

A very efficient Double Dummy Solver (DDS) has been written by Bo
Haglund \cite{haglund2010search}. In our experiments we use it to
evaluate double dummy hands. It makes use of partition
search \cite{ginsberg96} among many other optimizations to improve the
solving speed of the $\alpha\beta$ .

\subsection{Some Problems of PIMC}

PIMC is the state of the art of computer Bridge, it is used for
example in GIB \cite{ginsberg2001} and in
WBRIDGE5 \cite{ventos2017boosting} the current computer world
champion. The PIMC algorithm is given in algorithm \ref{PIMC}. In this
algorithm $S$ is the set of possible worlds and $allMoves$ is the set
of moves to be evaluated. The play function plays a move in a possible
world and returns the corresponding state. The doubleDummy function
evaluates the state using a double dummy solver.

There are multiple problems with
PIMC \cite{long2010understanding}. Here we will illustrates some
problems for the declarer with a No Trump contract since our
experiments use this restriction.

\begin{algorithm}
\begin{algorithmic}[1]
\STATE{Function $PIMC$ ($allMoves,S$)}
\begin{ALC@g}
\FOR{$move \in allMoves$}
\STATE{$score [move] \leftarrow 0$}
\FOR{$w \in S$}
\STATE{$s \leftarrow$ play ($move,w$)}
\STATE{$score [move] \leftarrow score [move] + $ doubleDummy ($s$)}
\ENDFOR
\ENDFOR
\RETURN $argmax_{move}(score [move])$
\end{ALC@g}
\end{algorithmic}
\caption{\label{PIMC}The PIMC algorithm.}
\end{algorithm}

\setboolean{hdsettings}{true}

\begin{handdiagram}
\north{KJT7,AKQ,AKQ,xxx}
\south{A986,xxx,xxx,AKQ}
\end{handdiagram}

In this hand from \cite{ginsberg2001}, PIMC finds that the declarer
always makes all of the four tricks at Spades. This problem is known
as strategy fusion \cite{frank1998search}. The reason why PIMC
misevaluates the hand is because it can play different cards in
different worlds whereas it should play the same cards in all the
worlds. Frank and Basin solve this problem with an algorithm they call
Vector Minimaxing \cite{frank2001theoretical} that plays the same
cards for the Max player in all the worlds. In Bridge terms the reason
why PIMC fails is that the finesse of the Queen of Spades always works
in all worlds given perfect information. This can be misleading for
the bidding phase also since Flat Monte Carlo thinks it can make four
tricks when it has only 50\% chances of making them thus reducing the
chances of making the contract to 50\% instead of 100\%. Another
problem with this hand is that the best way of playing is to play
other cards before trying the finesse since it can gain information on
the repartition of the Spades. If such information is available it is
best to try the finesse for the side that has the least Spades. This
is known as Discovery Play. The last thing about this hand is that if
South has decided to finesse the Queen of Spades at East, it should
first play the Ace in case the Queen is single.

\begin{handdiagram}
\north{J876,,,}
\south{AT32,,,}
\east{KQ5,,,}
\west{94,,,}
\end{handdiagram}

\begin{figure}
  \centering
  \caption{Example of a tree with three worlds illustrating non-locality.}
  \label{figureNonLocality}
\begin{tikzpicture}[level/.style={sibling distance=40mm/#1}]
\node [rectangle,draw] (z) {a}
  child {node [circle,draw,label=right:{$[0~0~0]$}] (b) {b} {
    child {node [rectangle,draw,,left=0.5cm,label=right:{$[0~1~1]$}] (d) {d}
      child {node (g) {$[1~0~0]$}}
      child {node (h) {$[0~1~1]$ }}
      }
    child {node [rectangle,draw,right=0.5cm,label=right:{$[1~0~0]$}] (d) {e}
      child {node (i) {$[0~0~0]$ }}
      child {node (j) {$[1~0~0]$ }}
    }
  }
  }
  child {node [circle,draw,label=right:{$[0~0~0]$}] (c) {c}
    child {node [rectangle,draw,label=right:{$[0~0~0]$}] (f) {f}
      child {node (g) {$[0~0~0]$ }}
    }
  };
\end{tikzpicture}
\end{figure}
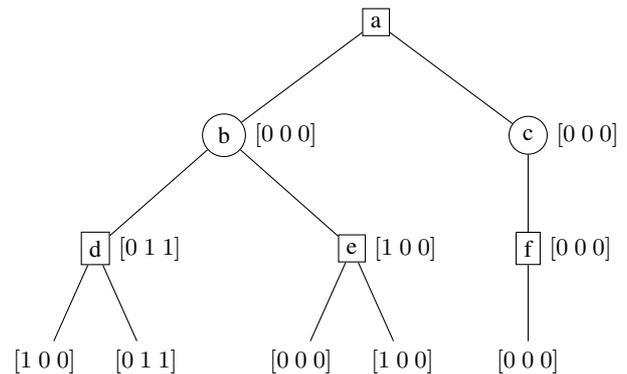

This hand from \cite{frank1998search} illustrates the problem of non
locality. In Bridge terms when the dummy plays the 6 East plays the
King and the declarer plays the Ace the best play is to finesse the 7
of Spades which is better than to finesse the Jack. From an
algorithmic point of view non locality can be explained using
figure \ref{figureNonLocality} from \cite{frank1998finding}. It
illustrates non-locality when searching with strategy fusion for Max
and perfect information for Min. As usual the Max nodes are squares
and the Min nodes are circles. The leaves gives the result of the game
in the three possible worlds. For example the move to the right from
node $d$ reaches a state labeled $[1~0~0]$ which means that the game
is won in world 1 (hence the 1 in the first position), lost in world 2
(hence the 0 in the second position) and also lost in world 3 (hence
the 0 in the third position). The vectors near the internal nodes give
the values that are backed up by the strategy fusion for Max and
perfect information for Min algorithm. We can see that each Max node
is evaluated by choosing the move that gives the maximum average
outcome. For example at node $d$ there are two moves, the left one
leads to $[1~0~0]$ and therefore has an average of $\frac{1}{3}$
whereas the right one leads to $[0~1~1]$ and has an average of
$\frac{2}{3}$. So node $d$ backs up $[0~1~1]$. However it is not
globally optimal. If instead of choosing the right move at node $d$ it
chooses the left move it backs up $[1~0~0]$ and then the $b$ node
would have been evaluated better also with $[1~0~0]$. It illustrates
that choosing the local optimum at node $d$ prevents from finding the
real optimum at node $b$. At Min nodes the algorithm chooses for each
world the minimal outcome over all children since it can choose the
move it prefers most in each different world.

\begin{handdiagram}
\north{xxx,,Kx,}
\south{K,,xxxx,}
\east{AQJ,,xx,}
\west{62,,Ax,x}
\end{handdiagram}

This hand illustrates that PIMC can drop the King of Spades as South
when West plays the small Clubs because the King is useless in DDS as
it is always taken by the Ace. In a real game the declarer can play a
small Spades from the dummy and it is not clear East will play the
Ace, leaving the possibility to win the trick with the King.

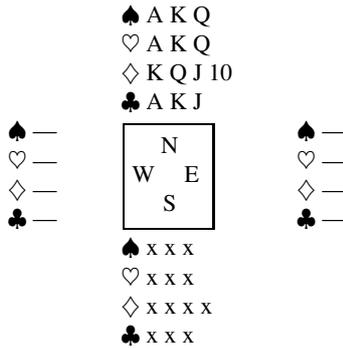
\begin{handdiagram}
\north{AKQ,AKQ,KQJT,AKJ}
\south{xxx,xxx,xxxx,xxx}
\east{,,,}
\west{,,,}
\end{handdiagram}

This hand comes from Fred Gitelman and illustrates reasoning on the
cards played by the defense to infer the possible hands. The contract
is 6 NT. If West starts with the Ace of Diamonds and then plays a
small Clubs the declarer can infer that West does not have the Queen
of Clubs and choose not to finesse the Queen. GIB for example does not
see it and finesse the Queen.

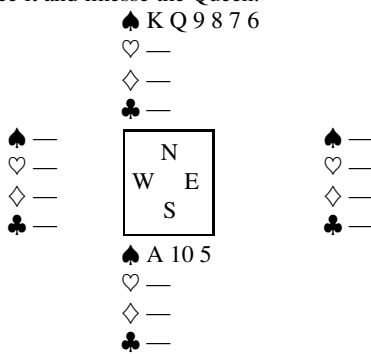
\begin{handdiagram}
\north{KQ9876,,,}
\south{AT5,,,}
\east{,,,}
\west{,,,}
\end{handdiagram}

This hand illustrates that strong Bridge players know how to play so
as to avoid rare negative events. Monte Carlo search does not see
these events if the corresponding worlds are not generated by the
sampling. The Spades are won 100\% of the time for a human player. The
only way to lose a trick at Spades is when the Jack is with the three
remaining Spades which is rare. A human player will play the King to
discover if it is the case and then play accordingly doing the right
finesse.

\section{Vectors of Outcomes and Pareto Fronts}

In this section we define Vectors and Pareto fronts that are used by
the algorithms in the next section.

\subsection{Definitions for Vectors}

Given $n$ different possible worlds, a vector of size $n$ keeps the
status of the game for each possible world. A zero at index n means
that the game is lost for world number n. A one means the game is
won. Associated to the vector there is another vector of booleans
indicating whether the world is possible in the current state. At the
root of the search all worlds are possible but when an opponent makes
a move, the move is usually only valid in some of the worlds and the
valid worlds are reduced.


The maximum of two vectors is a vector that for each index contains
the maximum of the two values at this index in the two
vectors. Similarly for the minimum.

A vector is greater or equal to another vector if for all indices it
contains a value greater or equal to the value contained at this index
in the other vector and if the valid worlds are the same for the two
vectors. A vector dominates another vector if it is greater or equal
to the other vector.

The score of a vector is the average among all possible worlds of the
values contained in the vector.

\subsection{Pareto Front}

A Pareto front is a set of vectors. It maintains the set of vectors
that are not dominated by other vectors. When a new vector is a
candidate for insertion in the front the first thing to verify is
whether the candidate vector is dominated by a vector in the front. If
it is the case the candidate vector is not inserted and the front
stays the same. If the candidate vector is not dominated it is
inserted in the front and all the vectors in the front that are
dominated by the candidate vector are removed.

For example consider the Pareto front $\{[1~0~0],[0~1~1]\}$. If we add
the vector $[0~0~1]$ to the front, then the front stays unchanged
since $[0~0~1]$ is dominated by $[0~1~1]$. If we add the vector
$[1~1~0]$ then the vector $[1~0~0]$ is removed from the front since it
is dominated by $[1~1~0]$, and then $[1~1~0]$ is inserted in the
front. The new front becomes $\{[1~1~0],[0~1~1]\}$.


It is useful to compare Pareto fronts. A Pareto front is greater or
equal to another Pareto front if for each element of the second Pareto
front there is an element in the first Pareto front which is greater
or equal to the element of the second Pareto front.



\section{Search with Strategy Fusion and Non-Locality}

In this section we explain in details the search algorithm and its
optimizations.

\subsection{Maximizing the probability of winning}

In Bridge the score of a board is calculated from the number of tricks
required by the contract and the number of won tricks. For instance,
let us consider the contract of 3NT used in our experiments where the
minimum number of tricks required is 9. For 9 won tricks, the score is
+400, for 10 won tricks the score is +430 since the bonus points for
one overtrick is only +30 but if the declarer gets only 8 tricks the
contract is defeated and the score is then -50. This threshold has an
impact on the card play, where the first goal is to make the contract
and then to try to obtain overtricks if it does not endanger the
contract. In our experiments we maximize the probability of making the
contract which is not optimal but reasonable.


\subsection{Search with Strategy Fusion}


Let assume that the defense know the cards of the declarer and that
the declarer optimizes against all possible states that corresponds to
his information. The score of a move for the declarer is the score of
the vector that has the best score among the vectors in the Pareto
front of the move. At a Max node the declarer computes after each move
the union of the Pareto fronts of all the moves that have been tried
so far. Min has knowledge of the declarer cards so in each world she
takes the move that minimizes the result of Max. The code for Min and
Max nodes is given in algorithm \ref{AlphaMu}. $\alpha\mu$ is a
generalization of PIMC since a search with a depth of one is PIMC.

The parameter $M$ controls the number of Max moves, when $M = 0$ the
algorithm reaches a leaf and each remaining possible world is
evaluated with a double dummy search. The stop function is given in
algorithm \ref{Terminal}. It also stops the search if the contract is
already won no matter what is played after. The parameter $state$
contains the current state where all the moves before have been played
and which does not contain the hidden information. The parameter
$Worlds$ contains the set of all possible worlds compatible with the
moves already played. The Pareto front is first initialized with an
empty set (line 5). If at a min node, the set of all possible moves in
all possible worlds is calculated (lines 7-11). For each move, the
move is played, the possible worlds updated and a recursive call is
performed. The Pareto front resulting from the recursive call is then
combined with the overall front (lines 12-17). We will explain later
the min algorithm. Similar operations are performed for a Max node
except that the combination with the overall front is then done with
the max algorithm (lines 19-28). We explain the max algorithm in the
next section.

\begin{algorithm}
\begin{algorithmic}[1]
\STATE{Function $\alpha\mu$ ($state,M,Worlds$)}
\begin{ALC@g}
\IF{$stop (state,M,Worlds,result)$}
\RETURN $result$
\ENDIF
\STATE{$front \leftarrow \emptyset$}
\IF{Min node}
\STATE{$allMoves \leftarrow \emptyset$}
\FOR{$w \in Worlds$}
\STATE{$l \leftarrow$ legalMoves ($w$)}
\STATE{$allMoves = allMoves \cup l$}
\ENDFOR
\FOR{$move \in allMoves$}
\STATE{$s \leftarrow$ play ($move,state$)}
\STATE{$W_1 \leftarrow \{w \in Worlds : move \in w\}$}
\STATE{$f \leftarrow \alpha\mu$ ($s,M,W_1$)}
\STATE{$front \leftarrow$ min($front,f$)}
\ENDFOR
\ELSE
\FOR{$w \in Worlds$}
\STATE{$l \leftarrow$ legalMoves ($w$)}
\STATE{$allMoves = allMoves \cup l$}
\ENDFOR
\FOR{$move \in allMoves$}
\STATE{$s \leftarrow$ play ($move,state$)}
\STATE{$W_1 \leftarrow \{w \in Worlds : move \in w\}$}
\STATE{$f \leftarrow \alpha\mu$ ($s,M-1,W_1$)}
\STATE{$front \leftarrow$ max($front,f$)}
\ENDFOR
\ENDIF
\RETURN $front$
\end{ALC@g}
\end{algorithmic}
\caption{\label{AlphaMu}The $\alpha\mu$ search algorithm without cuts and without transposition table.}
\end{algorithm}

\begin{algorithm}
\begin{algorithmic}[1]
\STATE{Function $stop$ ($state,M,Worlds,result$)}
\begin{ALC@g}
\IF{$declarerTricks(state) \geq contract$}
\FOR{$w \in Worlds$}
\STATE{$result [w] \leftarrow 1$}
\ENDFOR
\RETURN $True$
\ENDIF
\IF{$defenseTricks(state) > 13 - contract$}
\FOR{$w \in Worlds$}
\STATE{$result [w] \leftarrow 0$}
\ENDFOR
\RETURN $True$
\ENDIF
\IF{$M = 0$}
\FOR{$w \in Worlds$}
\STATE{$result [w] \leftarrow$ doubleDummy ($w$)}
\ENDFOR
\RETURN $True$
\ENDIF
\RETURN $False$
\end{ALC@g}
\end{algorithmic}
\caption{\label{Terminal}The function that stops search.}
\end{algorithm}

\subsection{Max nodes}

At Max nodes each possible move returns a Pareto front. The overall
Pareto front is the union of all the Pareto fronts of the moves. The
idea is to keep all the possible options for Max, i.e. Max has the
choice between all the vectors of the overall Pareto front. In order
to optimize computations and memory, vectors that are dominated by
another vector in the same Pareto front are removed.

\subsection{Min nodes}

The Min players can choose different moves in different possible
worlds. So they take the minimum outcome over all the possible moves
for a possible world. So when they can choose between two vectors they
take for each index the minimum between the two values at this index
of the two vectors.

Now when Min moves lead to Pareto fronts, the Max player can choose
any member of the Pareto front. For two possible moves of Min, the Max
player can also choose any combination of a vector in the Pareto front
of the first move and of a vector in the Pareto front of the second
move. In order to build the Pareto front at a Min node we therefore
have to compute all the combinations of the vectors in the Pareto
fronts of all the Min moves. For each combination the minimum outcome
is kept so as to produce a unique vector. Then this vector is inserted
in the Pareto front of the Min node.

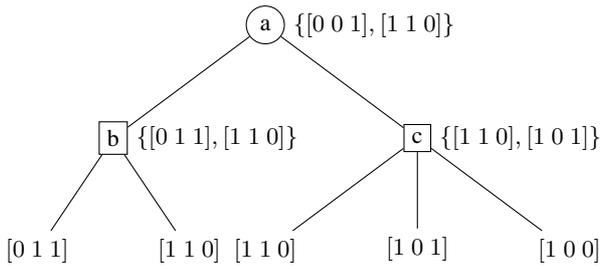
\begin{figure}
  \centering
  \caption{Product of Pareto fronts at Min nodes.}
  \label{figureProduct}
\begin{tikzpicture}[level/.style={sibling distance=40mm/#1}]
\node [circle,draw,label=right:{$\{[0~0~1],[1~1~0]\}$}] (z) {a}
  child {node [rectangle,draw,label=right:{$\{[0~1~1],[1~1~0]\}$}] (a) {b}
    child {node (b) {$[0~1~1]$ }}
    child {node (c) {$[1~1~0]$ }}
    }
  child {node [rectangle,draw,label=right:{$\{[1~1~0],[1~0~1]\}$}] (d) {c}
    child {node (e) {$[1~1~0]$ }}
     child {node [below=-0.3cm] (f) {$[1~0~1]$ }}
    child {node (g) {$[1~0~0]$ }}
    };
\end{tikzpicture}
\end{figure}

An example of the product of Pareto fronts is given in
figure \ref{figureProduct}. We can see in the figure that the left
move for Min at node $a$ leads to a Max node $b$ with two moves. The
Pareto front of this Max node is the union of the two vectors at the
leaves: $\{[0~1~1],[1~1~0]\}$. The right move for Min leads to a Max
node $c$ with three possible moves. When adding the vectors to the
Pareto front of the Max node $c$, the algorithm sees that $[1~0~0]$ is
dominated by $[1~0~1]$ and therefore does not add it to the Pareto
front at node $c$. So the resulting Pareto front for the Max node $c$
is $\{[1~1~0],[1~0~1]\}$. Now to compute the Pareto front for the root
Min node we perform the product of the two reduced Pareto fronts of
the children Max nodes and it gives:
$\{[0~1~0],[0~0~1],[1~1~0],[1~0~0]\}$. We then reduce the Pareto front
of the Min node and remove $[0~1~0]$ which is dominated by $[1~1~0]$
and also remove $[1~0~0]$ which is also dominated by
$[1~1~0]$. Therefore the resulting Pareto front for the root Min node
is $\{[0~0~1],[1~1~0]\}$.

We can also explain the behavior at Min nodes on the non-locality
example of figure \ref{figureNonLocality}. The Pareto front at Max
node $d$ is $\{[1~0~0],[0~1~1]\}$. The Pareto front at Max node $e$ is
$\{[0~0~0],[1~0~0]\}$. It is reduced to $\{[1~0~0]\}$ since $[0~0~0]$
is dominated. Now at node $b$ the product of the Pareto fronts at
nodes $d$ and $e$ gives $\{[1~0~0],[0~0~0]\}$ which is also reduced to
$\{[1~0~0]\}$. The Max player can now see that the $b$ node is better
than the $c$ node, it was not the case for the strategy fusion
algorithm without Pareto fronts.

The function to compute the minimum of two Pareto fronts is given in
algorithm \ref{Min}.


\begin{algorithm}
\begin{algorithmic}[1]
\STATE{Function $min$ ($front,f$)}
\begin{ALC@g}
\STATE{$result \leftarrow \emptyset$}
\FOR{$vector \in front$}
\FOR{$v \in f$}
\FOR{$w \in 0..size(vector)$}
\IF{$vector [w] < v [w]$}
\STATE{$r [w] \leftarrow vector [w]$}
\ELSE
\STATE{$r [w] \leftarrow v [w]$}
\ENDIF
\ENDFOR
\STATE{remove the vectors from $result \leq r$}
\IF{no vector from $result \geq r$}
\STATE{$result \leftarrow result \cup r$}
\ENDIF
\ENDFOR
\ENDFOR
\RETURN $result$
\end{ALC@g}
\end{algorithmic}
\caption{\label{Min}The function for joining two Pareto fronts at Min nodes.}
\end{algorithm}

\subsection{Skipping Min nodes}

The search one depth deeper at a Min node will always give the same
result as the Pareto front at that node since the Double Dummy Solver
has already searched all worlds with an $\alpha\beta$ and that the Min
player can choose the move for each world and therefore will have the
same result as the $\alpha\beta$ for each world.

This is why we only keep the number $M$ of Max moves to be played in
the search. The search will never stop after a Min move since
recursive calls at Min node do not decrease $M$. This is intended
since the results of the search after a Min move are the same as
before the Min move.

\subsection{Iterative Deepening and Transposition Table}

Iterative deepening starts with one Max move and increases the number
of Max moves at every iteration. The number of Max moves is the number
of Max nodes that have been traversed before reaching the current
state. The results of previous searches for all the nodes searched are
stored in a transposition table.

An entry in the transposition table contains the Pareto front of the
previous search at this node and the best move found by the search.


When a search is finished at a node, the entry in the transposition
table for this node is updated with the new Pareto front and the new
best move.







\subsection{Comparing Pareto Fronts}

A Pareto front $p_1$ is smaller or equal to another Pareto front $p$
if $p \cup p_1 = p$. When it is the case it is safe to ignore the move
associated to $p_1$ since it adds no options to $p$. If it is true for
the current front $p_1$ at a Min node it will also be true when
searching more this Min node since $p_1$ can only be reduced to a
smaller Pareto front by more search at a Min node.

An efficient way to compare $p$ to $p_1$ is to ensure that each vector
of $p_1$ is dominated by another vector in $p$. The corresponding
algorithm is given in algorithm \ref{Smaller}.

\begin{algorithm}
\begin{algorithmic}[1]
\STATE{Function $\leq$ ($front,f$)}
\begin{ALC@g}
\FOR{$vector \in front$}
\STATE{$oneGreaterOrEqual \leftarrow False$}
\FOR{$v \in f$}
\IF{$vector <= v$}
\STATE{$oneGreaterOrEqual \leftarrow True$}
\STATE{$break$}
\ENDIF
\ENDFOR
\IF{$oneGreaterOrEqual = False$}
\RETURN $False$
\ENDIF
\ENDFOR
\RETURN $True$
\end{ALC@g}
\end{algorithmic}
\caption{\label{Smaller}The function to test if a Pareto front is smaller than another one.}
\end{algorithm}


\subsection{Early Cut}



If a Pareto front at a Min node is dominated by the Pareto front of
the upper Max node it can safely be cut since the evaluation is
optimistic for the Max player. The Max player cannot get a better
evaluation by searching more under the Min node and it will always be
cut whatever the search below the node returns since the search below
will return a Pareto front smaller or equal to the current Pareto
front. It comes from the observation that a world lost at a node is
also lost at all nodes below.

\begin{figure}
  \centering
  \caption{Example of an early cut at node c.}
  \label{figureEarlyCut}
\begin{tikzpicture}[level/.style={sibling distance=40mm/#1}]
\node [rectangle,draw] (z) {a}
  child {node [circle,draw,label=right:{$\{[1~1~0],[0~1~1]\}$}] (b) {b} 
     child {node (d) {$[1~1~0]$ }}
     child {node (e) {$[0~1~1]$}}
  }
  child {node [circle,draw,label=right:{$[1~1~0] \rightarrow cut$}] (c) {c}
  };
\end{tikzpicture}
\end{figure}
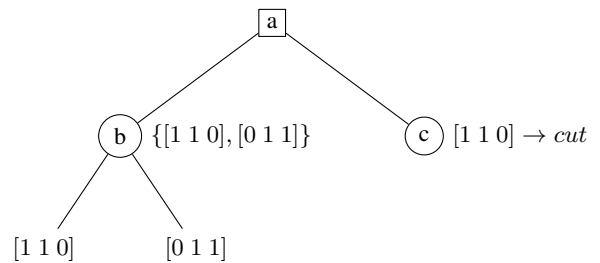

Figure \ref{figureEarlyCut} gives an example of an early cut at a Min
node. The root node $a$ is a Max node, the first move played at $a$
returned $\{[1~1~0],[0~1~1]\}$. The second move is then tried leading
to node $c$ and the initial Pareto front calculated with double dummy
searches at node $c$ is [1 1 0]. It is dominated by the Pareto front
of node $a$ so node $c$ can be cut.

\subsection{Root Cut}

If a move at the root of $\alpha\mu$ for $M$ Max moves gives the same
probability of winning than the best move of the previous iteration of
iterative deepening for $M-1$ Max moves, the search can be safely be
stopped since it is not possible to find a better move. A deeper
search will always return a worse probability than the previous search
because of strategy fusion. Therefore if the probability is equal to
the one of the best move of the previous shallower search the
probability cannot be improved and a better move cannot be found so it
is safe to cut.

\subsection{$\alpha\mu$}


$\alpha\mu$ with transposition table and cuts is a search algorithm
using Pareto fronts as evaluations and bounds. The algorithm is given
in algorithm \ref{AlphaMuTT}.

The evaluation of a state at a leaf node is the double dummy
evaluation for each possible world. An evaluation for a world is 0 if
the game is lost for the Max player and 1 if the game is won for the
Max player (lines 2-5).

The algorithm starts with getting the entry $t$ of $state$ in the
transposition table (line 6). The entry contains the last Pareto front
found for this state and the best move found for this state, i.e. the
move associated to the best average.

If the state is associated to a Min node, i.e. a Min player is to
play, the algorithm starts to get the previously calculated Pareto
front from the transposition table (line 8). Then it looks for an
early cut (lines 9-11). If the node is not cut it computes the set of
all possible moves over all the valid worlds (lines 12-16). It then
moves the move of the transposition table in front of the possible
moves (line 17). After that it tries all possible moves (line 18). For
each possible move it computes the set $W_1$ of worlds still valid
after the move and recursively calls $\alpha\mu$ (lines 19-21) . The
parameters of the recursive call are $s$, the current state, $M$ the
number of Max moves to go which is unchanged since we just played a
Min move, $W_1$ the set of valid worlds after $move$, and an empty set
for $alpha$ to avoid deeper cuts. The front returned by the recursive
call is then combined to the current front using the min function
(line 22). When the search is finished it updates the transposition
table and returns the $mini$ Pareto front (lines 24-25).

If the state is associated to a Max node it initializes the resulting
front with an empty set (line 27). Then as in the Min nodes it
computes the set of all possible moves and moves the transposition
table move in front of all the possible moves (lines 28-32). Then it
tries all the moves and for each move computes the new set $W_1$ of
valid worlds and recursively calls $\alpha\mu$ with $M-1$ since a Max
move has just been played and $front$ as $alpha$ since a cut can
happen below when the move does not improve $front$ (lines 33-36). The
resulting front $f$ is combined with front with the max function (line
37). If the score of the best move ($\mu(front)$) is equal to the
score of the best move of the previous search and the node is the root
node then a Root cut is performed (lines 38-42). When the search is
finished the transposition table is updated and $front$ is returned
(lines 44-45).

The search with strategy fusion is always more difficult for the Max
player than the double dummy search where the Max player can choose
different moves in the different possible worlds for the same
state. Therefore if a double dummy search returns a loss in a possible
world, it is sure that the search with $\alpha\mu$ will also return a
loss for this world.

If the search is performed until terminal nodes and all possible
worlds are considered then $\alpha\mu$ solves the strategy fusion and
the non locality problem for the game where the defense has perfect
information.

If the search is stopped before terminal nodes and not all possible
worlds are considered then $\alpha\mu$ is a heuristic search
algorithm.

The algorithm is named $\alpha\mu$ since it maximizes the mean and
uses an $\alpha$ bound.

\begin{algorithm}
\begin{algorithmic}[1]
\STATE{Function $\alpha\mu$ ($state,M,Worlds,\alpha$)}
\begin{ALC@g}
\IF{$stop (state,M,Worlds,result)$}
\STATE{update the transposition table}
\RETURN $result$
\ENDIF
\STATE{$t \leftarrow$ entry in the transposition table}
\IF{Min node}
\STATE{$mini \leftarrow \emptyset$}
\IF{$t.front \leq \alpha$}
\RETURN $mini$
\ENDIF
\STATE{$allMoves \leftarrow \emptyset$}
\FOR{$w \in Worlds$}
\STATE{$l \leftarrow$ legalMoves ($w$)}
\STATE{$allMoves = allMoves \cup l$}
\ENDFOR
\STATE{move $t.move$ in front of $allMoves$}
\FOR{$move \in allMoves$}
\STATE{$s \leftarrow$ play ($move,state$)}
\STATE{$W_1 \leftarrow \{w \in Worlds : move \in w\}$}
\STATE{$f \leftarrow \alpha\mu$ ($s,M,W_1,\emptyset$)}
\STATE{$mini \leftarrow$ min($mini,f$)}
\ENDFOR
\STATE{update the transposition table}
\RETURN $mini$
\ELSE
\STATE{$front \leftarrow \emptyset$}
\FOR{$w \in Worlds$}
\STATE{$l \leftarrow$ legalMoves ($w$)}
\STATE{$allMoves = allMoves \cup l$}
\ENDFOR
\STATE{move $t.move$ in front of $allMoves$}
\FOR{$move \in allMoves$}
\STATE{$s \leftarrow$ play ($move,state$)}
\STATE{$W_1 \leftarrow \{w \in Worlds : move \in w\}$}
\STATE{$f \leftarrow \alpha\mu$ ($s,M-1,W_1,front$)}
\STATE{$front \leftarrow$ max($front,f$)}
\IF{root node}
\IF{$\mu(front) = \mu$ of previous search}
\STATE{break}
\ENDIF
\ENDIF
\ENDFOR
\STATE{update the transposition table}
\RETURN $front$
\ENDIF
\end{ALC@g}
\end{algorithmic}
\caption{\label{AlphaMuTT}The $\alpha\mu$ search algorithm with cuts and transposition table.}
\end{algorithm}


\subsection{Equivalent Cards and Partitions}

Before performing the Iterative Deepening search the program checks
whether it is useful to perform a search. For example if there is only
one possible move which often happens or if there are two equivalent
moves. To detect that two moves of the same color are equivalent the
program normalizes the state using the same idea as in Partition
Search \cite{ginsberg96}. In the normalized state all the cards in a
color have consecutive values. It is then easy to detect that two
cards in the same hand are equivalent: they have consecutive values.

\subsection{Generating Possible Worlds}

Before performing the search the program generates the set of possible
worlds. The principle of the generation is to randomly generate worlds
and to retain those that satisfy a set of constraints. The constraints
are constraints on the initial deal corresponding to the generated
world. The reconstructed initial deal must comply with the constraints
on the contract. The generated world also has to comply with the known
sluffs of the other players.

\section{Experimental Results}

In our experiments we fix the bids so as to concentrate on the
evaluation of the card play. We use the one no trump, pass, three no
trump, pass, pass, pass bid for all experiments.

We use duplicate scoring. It means that the different evaluated
programs will play the same hands against the same opponents. When
$\alpha\mu$ is the declarer it will play against two PIMC as the
defense. $\alpha\mu$ is a generalization of PIMC since $\alpha\mu$
at depth one is PIMC. So in order to compare $\alpha\mu$ as a declarer
to PIMC as a declarer we compare $\alpha\mu$ as a declarer to
$\alpha\mu$ with $M=1$ as a declarer.

There are constraints on the hands due to the contract. Initial deals
and possible worlds for PIMC and $\alpha\mu$ are generated according
to the constraints. However when using no more constraints many
initial deals are useless for evaluating the program since they are
always won or always lost and that they do not discriminate between
programs since all the programs have the same result. In order to
alleviate this problem we only keep initial deals where PIMC has more
than 30\% and less than 70\% winning rate, i.e. the undecided and
balanced deals.

We first test $\alpha\mu$ at different depth versus PIMC with a fixed
number of possible worlds. Table \ref{table13cardsScore} gives the
results for different runs of $\alpha\mu$ as the declarer with 20
worlds and 40 worlds against PIMC as the defense with 20 worlds. All
results are computed playing the same 500 initial deals with the same
seed for each deal. We see that looking two or three Max moves ahead
can be beneficial. The number of discrepancies is the number of times
$\alpha\mu$ chooses a different move than the move at depth one
(i.e. the PIMC move). We note that it happens relatively rarely. PIMC
is already a very strong player as a declarer so improving on it even
slightly is difficult. For 52 cards and 20 worlds PIMC ($\alpha\mu$
with $M=1$) scores 60.2\% and $\alpha\mu$ with $M=3$ scores
62.0\%. For 52 cards and 40 worlds PIMC ($\alpha\mu$ with $M=1$)
scores 62.4\% and $\alpha\mu$ with $M=3$ scores 63.2\%. For 36 cards
and 20 worlds PIMC ($\alpha\mu$ with $M=1$) scores 46.4\% and
$\alpha\mu$ with $M=3$ scores 48.2\%. We can conclude that $\alpha\mu$
improves on PIMC.

\begin{table}
  \centering
  \caption{Comparison of the scores of different configurations of
  $\alpha\mu$ on deals with 52 or 36 cards.}
  \label{table13cardsScore}
  \begin{tabular}{rrrrrrrrrrr}
Cards & M & Worlds & Discrepancies &  Score \\
      &   &        &               &        \\
   52 & 1 &     20 &    0 / 13 000 & 60.2\% \\
   52 & 2 &     20 &  169 / 13 000 & 63.0\% \\
   52 & 3 &     20 &  276 / 13 000 & 62.0\% \\
   52 & 1 &     40 &    0 / 13 000 & 62.4\% \\
   52 & 2 &     40 &  213 / 13 000 & 62.4\% \\
   52 & 3 &     40 &  388 / 13 000 & 63.2\% \\
   36 & 1 &     20 &    0 / 13 000 & 46.4\% \\
   36 & 2 &     20 &  124 / 13 000 & 47.8\% \\
   36 & 3 &     20 &  190 / 13 000 & 48.2\% \\
  \end{tabular}
\end{table}

We now compare the times to play moves with and without Transposition
Tables and cuts. Table \ref{table13cardsTime} gives the average time
per move of different configurations of $\alpha\mu$ playing entire
games. TT means Transposition Table, R means Root Cut, E means Early
Cut. We can observe that a Transposition Table associated to cuts
improves the search time. For $M=1$ the search time is 0.096
seconds. For $M=3$ without transposition table and cuts the average
search time per move is 18.678 seconds. When using a transposition
table associated to early and root cuts it goes down to 1.228 seconds.

\begin{table}
  \centering
  \caption{Comparison of the average time per move of different
  configurations of $\alpha\mu$ on deals with 52 or 36 cards.}
  \label{table13cardsTime}
  \begin{tabular}{rrrrrrrrrrr}
Cards & M & Worlds & TT & R & E &   Time\\
      &   &        &    &   &   &       \\
   52 & 1 &     20 &    &   &   &  0.096\\
   52 & 2 &     20 &  n & n & n &  1.306\\
   52 & 2 &     20 &  y & y & n &  0.389\\
   52 & 2 &     20 &  y & n & y &  0.436\\
   52 & 2 &     20 &  y & y & y &  0.363\\
   52 & 3 &     20 &  n & n & n & 18.678\\
   52 & 3 &     20 &  y & y & n &  4.089\\
   52 & 3 &     20 &  y & n & y &  1.907\\
   52 & 3 &     20 &  y & y & y &  1.228\\
  \end{tabular}
\end{table}

\section{Conclusion and Future Work}

We presented the $\alpha\mu$ algorithm for Bridge card play. It
assumes the opponents have perfect information. It enables to search a
few moves ahead taking into account the strategy fusion and the non
locality problems. To solve the non locality problem it uses Pareto
fronts as evaluations of states and combines them in an original way
at Min and Max nodes. To solve the strategy fusion problem it plays
the same moves in all the valid worlds during search. Experimental
results for the 3NT contract shows it improves on PIMC.

We also presented the use of a transposition table as well as the
early and the root cut for $\alpha\mu$. When searching three Max moves
ahead it enables the search to be fifteen times faster while returning
the same move as the longer search without the optimizations.


In future work we expect to use partition Search with $\alpha\mu$. We
also plan to take into account that the defense only has incomplete
information and to take advantage of that for the declarer. We will
also apply the algorithm to the defense play. It should also improve
the algorithm to deal with real scores instead of only win/loss.

\section*{Acknowledgment}

Thanks to Alexis Rimbaud for explaining me how to use the solver of Bo
Haglund and to Bo Haglund for his Double Dummy Solver.

\bibliographystyle{ecai}
\bibliography{nonlocality}

\end{document}